\documentclass[12pt]{article}
\pdfoutput=1

\usepackage[round]{natbib}

\usepackage{hyperref}       
\usepackage{url}            
\usepackage{booktabs,siunitx}       
\usepackage{amsfonts}       
\usepackage{nicefrac}       
\usepackage{microtype}      
\usepackage{cite}
\usepackage{amsmath,amssymb,amsfonts}
\usepackage{algorithmic}
\usepackage{graphicx}
\usepackage{textcomp}
\usepackage{float}
\usepackage{array,multirow,graphicx}
\usepackage{animate}
\usepackage{arydshln}
\usepackage{mathtools}
\usepackage{xmpmulti}
\usepackage{alltt}
\usepackage{makecell}
\usepackage{hyperref}
\usepackage{bm}
\usepackage[dvipsnames]{xcolor}
\definecolor{ForestGreen}{RGB}{11, 148, 43}
\definecolor{magenta}{RGB}{247,29,176}
\definecolor{orange}{RGB}{219,106,72}
\definecolor{skyblue}{RGB}{42,138,153}
\definecolor{purple}{RGB}{153,51,255}
\usepackage{easybmat}
\usepackage{soul}
\usepackage[ruled, vlined]{algorithm2e}
\usepackage{enumitem}
\usepackage{algorithmic}

\newcommand{\vect}[1]{\boldsymbol{\mathbf{#1}}}

\DeclarePairedDelimiter\ceil{\lceil}{\rceil}

\newcommand{\convp}{\overset{P}{\to}}

\makeatletter
\renewcommand*{\@fnsymbol}[1]{\@alph{#1}}
\makeatother

\usepackage[margin=1in]{geometry}
\usepackage{footnote}
\usepackage[utf8]{inputenc}

\usepackage{eqparbox,collcell}
\usepackage{dsfont}

\usepackage{setspace}

\newcommand{\columnalign}{}
\newcommand{\columnalignment}[2]{\renewcommand{\columnalign}{\eqmakebox[#1][#2]}}
\usepackage{xcolor}
\usepackage{dblfloatfix}
\newcommand\numberthis{\addtocounter{equation}{1}\tag{\theequation}}

\begin{document}

%

%

\title{\bf Feature Selection for Huge Data via Minipatch Learning}
\author{Tianyi Yao\thanks{Department of Statistics, Rice University, Houston, TX}\hspace{.2cm} and Genevera I. Allen\thanks{Departments of Electrical and Computer Engineering, Statistics, and Computer Science, Rice University, Houston, TX} \textsuperscript{,}\thanks{Jan and Dan Duncan Neurological Research Institute, Baylor College of Medicine, Houston, TX}}
\date{}
\maketitle

\begin{abstract}
\noindent Feature selection often leads to increased model interpretability, faster computation, and improved model performance by discarding irrelevant or redundant features. While feature selection is a well-studied problem with many widely-used techniques, there are typically two key challenges: i) many existing approaches become computationally intractable in huge-data settings with millions of observations and features; and ii) the statistical accuracy of selected features degrades in high-noise, high-correlation settings, thus hindering reliable model interpretation. We tackle these problems by proposing Stable Minipatch Selection (STAMPS) and Adaptive STAMPS (AdaSTAMPS).  These are meta-algorithms that build ensembles of selection events of base feature selectors trained on many tiny, (adaptively-chosen) random subsets of both the observations and features of the data, which we call minipatches. Our approaches are general and can be employed with a variety of existing feature selection strategies and machine learning techniques. In addition, we provide theoretical insights on STAMPS and empirically demonstrate that our approaches, especially AdaSTAMPS, dominate competing methods in terms of feature selection accuracy and computational time.
\end{abstract}


\section{Introduction}

Feature selection is critical in statistical machine learning for improving interpretability of machine learning models. While feature selection has been used in a wide variety of applications, there are typically two key challenges: i) many existing techniques can quickly become computationally intractable in huge-data settings on the order of millions of observations and features; and ii) the statistical accuracy of selected features degrades in the presence of noise and/or highly correlated features, thus hampering reliable model interpretation.

\vspace{0.5em}
In this work, our primary goal is to develop practical feature selection frameworks that can select statistically accurate features in a computationally efficient manner, even with huge data in high-noise, high-correlation settings. We place great emphasis on selecting the correct features, which is a statistical accuracy problem, because doing so would considerably enhance explainability of machine learning methods, enable practitioners to reliably interpret models, and provide insights into the underlying data-generating processes (e.g. candidate gene study \citep{candidategenestudy}). As stressed in \citet{Lipton2018TheMO}, interpretability of models is critical in many machine learning applications. Furthermore, we seek to design our feature selection framework to be conducive to efficient computation so that it can be applied in huge-data settings with millions of features and observations. Data of such scale is regularly observed in online marketing, genetics, neuroscience, and text mining, among many other areas of research.

\vspace{0.5em}
Popular feature selection methods can be divided into three categories - filter, wrapper, and embedded methods \citep{guyon2003jmlr}. Filter methods select features based solely on observed characteristics of available data independently of a learning algorithm \citep{Jovic2015ARO}. While filter methods are fast computationally and they might work well for prediction by selecting features related to the output, filter methods usually perform poorly in terms of statistical accuracy of selecting the right features that are critical for reliable model interpretation because they tend to select sets of correlated, redundant features together.

\vspace{0.5em}
Unlike filter methods, wrapper methods assess the quality of selected features using performance of a predefined learning algorithm. Therefore, wrappers are much slower computationally than filter methods \citep{Jovic2015ARO, pirgazi2019}. Despite generally better performance than filter methods, wrappers tend to perform suboptimally in selecting statistically accurate features because they are inherently greedy methods.

\vspace{0.5em}
Another line of work focuses on embedded methods, which can be further divided into optimization-based methods and methods that select features based on properties of the learner. The Lasso \citep{tibshirani1996lasso} is arguably one of the most widely-used optimization-based feature selection methods. The Lasso tends to do better on statistical accuracy of selected features and is model selection consistent under theoretical conditions including the Irrepresentable Condition \citep{zhao2006consistency, meinshausen2006neighborhood}. Yet, such theoretical conditions requiring low correlation among features will be less likely to ever hold in high-dimensional settings in which features tend to become increasingly correlated. Additionally, choosing the right amount of regularization for accurate feature selection is challenging in practice \citep{meinshausen2010stability}. Fitting the Lasso with model selection procedures (e.g. cross-validation) can be computationally challenging in huge-data settings.

\vspace{0.5em}
Last but not least, various tree-based algorithms such as Random Forest (RF) \citep{breiman2001} constitute a great proportion of embedded methods that select features using properties of the learner. In particular, numerous importance scores such as the RF permutation importance are used to rank features based on their contributions to improving predictive performance of the learner. However, previous studies have found many such importance measures can be biased and even become unstable in high-dimensional, high-correlation settings \citep{strobl2007, nicodemus2007impact, GENUER20102225}, thus hindering reliable model interpretation. Additionally, features selected using properties of the learner are often sensitive to the choice of tuning hyperparameters, and systematic hyperparameter search poses great computational challenges in huge-data settings.

\vspace{0.5em}
The idea of randomly subsampling observations and/or features for model training has appeared in many parts of machine learning, including various ensemble learning techniques
\citep{breiman1996, breiman2001, FREUND1997119, loupperandompatch, gomes2019stream, LeJeune2020TheIR} and the dropout technique \citep{JMLR:v15:srivastava14a} combined with stochastic gradient methods \citep{hardt2016sgd} in training deep neural networks. Additionally, there is another line of work focusing on using various data randomization techniques to select more ``stable'' features for high-dimensional linear regression \citep{bach2008, meinshausen2010stability, wang2011, shah2013cpss, yu2013, beinrucker2016stat, staerk2019highdimensional}.
Inspired by these approaches, we propose general and practical strategies that exploit tiny, (adaptively-chosen) random subsets of both the observations and features of the data for considerably enhancing feature selection accuracy and computational efficiency, hence improving model interpretability.


\paragraph{Contributions} We summarize our contributions as follows: We develop practical feature selection methods named STAMPS and AdaSTAMPS that leverage random data perturbation as well as adaptive feature sampling schemes to select statistically accurate features with dramatically improved robustness and computational efficiency (see Sec. 2). Our proposed approaches are general meta-algorithms that can be employed with a variety of existing feature selection strategies and machine learning techniques in practice. In addition, we theoretically show that STAMPS achieves model-agnostic familywise error rate (FWER) control under certain conditions. We empirically demonstrate the practical effectiveness of our approaches on multiple synthetic and real-world data sets in Sec. 3, showing that our frameworks, especially AdaSTAMPS, dominate competing methods in terms of both feature selection accuracy and computational time.

\section{Minipatch Feature Selection}

\subsection{Minipatch Learning}

People have used random subsets of observations and features for model training in many parts of machine learning (see Sec. 1). Notably in the ensemble learning literature, some have called these ``random patches'' \citep{loupperandompatch, gomes2019stream}. We call tiny, (adaptively-chosen) random subsets of both observations and features ``minipatches'' to denote the connection with minibatches as well as patches in image processing, and to also denote that these are typically very tiny subsets. Using tiny minipatches can yield major computational advantages for huge data. Formally, given a pair of response vector $\mathbf{y}\in\mathbb{R}^N$ and data matrix $\mathbf{X}\in\mathbb{R}^{N\times M}$ that consists of $N$ observations each having $M$ features, a minipatch can be obtained by simultaneously subsampling $n$ rows (observations) and $m$ columns (features) without replacement using some form of randomization, with $n \ll N$ and $m \ll M$.

\subsection{Stable Minipatch Selection (STAMPS)}


Inspired by stability selection \citep{meinshausen2010stability} and its extensions \citep{shah2013cpss, beinrucker2016stat} that keep ``stable'' features based on selection frequencies, we propose a general and practical approach to stable feature selection named Stable Minipatch Selection (STAMPS). STAMPS is a flexible meta-algorithm that can be employed with a variety of existing feature selection strategies and machine learning techniques.

\vspace{0.5em}
Our proposed STAMPS method is summarized in Algorithm \ref{alg:stamps}. Here, $\mathbf{y}_{I_k}$ denotes the subvector of $\mathbf{y}$ containing its elements indexed by $I_k$. Similarly, $\mathbf{X}_{I_k,F_k}$ denotes the submatrix of $\mathbf{X}$ containing its rows indexed by $I_k$ and its columns indexed by $F_k$. For brevity, $[M]$ denotes the set $\{1,2,\hdots,M\}$.

\vspace{0.5em}
Specifically, STAMPS fits arbitrary base feature selectors to many tiny, random minipatches and calculates feature selection frequencies by taking an ensemble of estimated feature supports over these minipatches. We define the selection frequency of the $j$\textsuperscript{th} feature at the $k$\textsuperscript{th} iteration, $\hat{\Pi}_j^{(k)}$, to be the number of times it is sampled and then selected by base feature selectors divided by the number of times it is sampled into minipatches after $k$ iterations. STAMPS finally outputs a set of stable features $\hat{S}^{\text{stable}}$ whose selection frequencies are above a user-specific threshold $\pi_{\text{thr}}\in(0,1)$. We discuss the choice of $\pi_{\text{thr}}$ in Sec. 2.5.

{\centering
\begin{algorithm}
\label{alg:stamps}
\caption{STAMPS }
\SetAlgoLined
\KwIn{$\mathbf{y}\in\mathbb{R}^{N}$, $\mathbf{X}\in\mathbb{R}^{N\times M}$, $n$, $m$, $\pi_{\text{thr}}\in(0,1)$.}
{\bf Initialize:} $\hat{\Pi}_j^{(0)}=0$, $\forall j\in [M]$\;
\While{stopping criterion not met}{
    \vspace{4pt}
    1) Sample a minipatch: subsample $n$ observations $I_k\subset [N]$ and $m$ features $F_k\subset [M]$ uniformly at random without replacement to get a minipatch $(\mathbf{y}_{I_k}, \mathbf{X}_{I_k,F_k})\in\mathbb{R}^n\times\mathbb{R}^{n\times m}$\;
    \vspace{4pt}
    2) Fit a selector to minipatch $(\mathbf{y}_{I_k}, \mathbf{X}_{I_k,F_k})$ to obtain an estimated feature support $\hat{\mathcal{S}}_k\subseteq F_k$\;
    \vspace{4pt}
    3) Ensemble feature supports $\{\hat{\mathcal{S}}_i\}_{i=1}^k$ to update selection frequencies $\hat{\Pi}_j^{(k)}, \forall j \in [M]$:
    \scalebox{0.9}{\parbox{1\linewidth}{
    \vspace{-8pt}
    \begin{align*}
        \hat{\Pi}^{(k)}_j = \frac{1}{\text{max}(1,\sum_{i=1}^k \mathds{1}(j\in F_i))}\sum_{i=1}^k\mathds{1}(j\in F_i, j\in \hat{\mathcal{S}}_i);
    \end{align*}
    }}
\vspace{-13pt}
}
\KwOut{$\hat{S}^{\text{stable}} = \big\{j\in[M]: \hat{\Pi}^{(K)}_j \geq \pi_{\text{thr}} \big\}$.}
\end{algorithm}
}

To avoid a forever STAMPS situation, one should employ some type of stopping criterion for Algorithm \ref{alg:stamps}. While there are many possible choices, we happen to use a simple one that is effective in our empirical studies - stop the algorithm if the rank ordering of the top $\text{min}(\text{max}(|\mathcal{H}|, \tau_{\text{l}}), \tau_{\text{u}})$ features in terms of selection frequencies $\{\hat{\Pi}_j^{(k)}\}_{j=1}^M$ remain unchanged for the past $100$ iterations, where $\mathcal{H}=\{j:\hat{\Pi}^{(k)}_j\geq 0.5\}$. Here, $\tau_{\text{l}}$ and $\tau_{\text{u}}$ are fixed, user-specific parameters. We suggest setting both to well exceed the approximate number of true signal features, with $\tau_{\text{u}}$ being an integer multiple of $\tau_{\text{l}}$.

\subsection{Adaptive Stable Minipatch Selection (AdaSTAMPS)}

We propose to adaptively sample minipatches in ways that further boost computational efficiency as well as feature selection accuracy. Replacing uniform feature sampling in step 1 of Algorithm \ref{alg:stamps} with any appropriate adaptive feature sampling scheme, we obtain our Adaptive STAMPS (AdaSTAMPS) framework for feature selection.

\vspace{0.5em}
While there are many possible adaptive sampling techniques to choose from, we develop two types of example algorithms in this work. We describe an exploitation and exploration scheme inspired by the multi-armed bandit literature \citep{Bouneffouf2019ASO, Slivkins2019IntroductionTM} here in Algorithm \ref{alg:adastamps}. For brevity, we use the shorthand AdaSTAMPS (EE) to denote the specific instance of AdaSTAMPS that employs Algorithm \ref{alg:adastamps} as its adaptive feature sampling scheme. In addition, we give another probabilistic adaptive feature sampling scheme in the Appendix, which we abbreviate as AdaSTAMPS (Prob). We use these two in our empirical studies just to show examples of what can be done with our flexible AdaSTAMPS approach. However, one can use many other more sophisticated sampling techniques in practice. We save the investigation of various other adaptive sampling schemes for future work.

\vspace{0.5em}
In Algorithm \ref{alg:adastamps}, $k$ is the iteration counter, $E$ denotes the total number of burn-in epochs, $\{\gamma^{(k)}\}$ is an increasing geometric sequence in the range $[0.5,1]$ for controlling the trade-off between exploitation and exploration of the input feature space, and $\pi_{\text{active}}\in(0,1)$ is a fixed, user-specific threshold for determining the active feature set. We found setting $\pi_{\text{active}}=0.1$ works well for most problems.

{\centering
\begin{algorithm}[h]
\label{alg:adastamps}
\caption{Adaptive Feature Sampling Scheme: Exploitation and Exploration (EE)}
\SetAlgoLined
\KwIn{$k$, $M$, $m$, $E$, $\{\hat{\Pi}_j^{(k-1)}\}_{j=1}^M$, $\{\gamma^{(k)}\}=\text{geom\_seq}([0.5, 1])$, $\pi_{\text{active}}$.}
\vspace{3pt}
{\bf Initialize:} $G=\ceil{\frac{M}{m}}$, $\mathcal{J}=\{1,\hdots,M\}$\;
\vspace{3pt}
    \uIf(\tcp*[h]{Burn-in stage}){$k \leq E\cdot G$}{
        \vspace{5pt}
        \uIf(\tcp*[h]{A new epoch}){$\text{mod}_{G}(k)=1$}{
            \vspace{3pt}
            1) Randomly reshuffle feature index set $\mathcal{J}$ and partition into disjoint sets $\{\mathcal{J}_g\}_{g=0}^{G-1}$\;
        }
        \vspace{3pt}
        1) Set $F_k=\mathcal{J}_{\text{mod}_{G}(k)}$\;
    }
    \Else(\tcp*[h]{Adaptive stage}){
        \vspace{3pt}
        1) Update active feature set: $\mathcal{A}=\big\{j\in\{1,\hdots,M\}:\hat{\Pi}^{(k-1)}_j\geq \pi_{\text{active}}\big\}$\;
        \vspace{3pt}
        2) Exploitation step: Sample $\text{min}(m, \gamma^{(k)}|\mathcal{A}|)$ features $F_{k,1}\subseteq \mathcal{A}$ uniformly at random\;
        \vspace{3pt}
        3) Exploration step: Sample $(m-\text{min}(m, \gamma^{(k)}|\mathcal{A}|))$ features $F_{k,2}\subseteq \{1,\hdots,M\}\setminus\mathcal{A}$ uniformly at random\;
        \vspace{3pt}
        4) Set $F_k = F_{k,1} \cup F_{k,2}$\;

    }

{\bf Output:} $F_k$.
\end{algorithm}
}

During the initial burn-in stage of Algorithm \ref{alg:adastamps},  AdaSTAMPS (EE) explores the entire input feature space by ensuring each feature is sampled into minipatches for exactly $E$ times, once for each epoch. The selection frequency of each feature is continuously updated as feature selection is repeatedly conducted on each random minipatch. After $E$ epochs, Algorithm \ref{alg:adastamps} transitions to the adaptive stage during which it exploits a subset of features that are likely to be true signal features while adaptively exploring the rest of the feature space. Specifically, features with selection frequencies above $\pi_{\text{active}}$ are kept in the active feature set $\mathcal{A}$, which is likely a superset of most of the true signal features. As $k$ increases and feature selection frequencies keep polarizing, $\mathcal{A}$ continues to shrink and hone in on the set of true signal features. To exploit the relatively stable features in $\mathcal{A}$, Algorithm \ref{alg:adastamps} samples an increasing proportion of $\mathcal{A}$ into each new minipatch by gradually increasing $\gamma^{(k)}$. Meanwhile, Algorithm \ref{alg:adastamps} continues to explore the rest of the feature space by drawing from $\{1,\hdots,M\}\setminus\mathcal{A}$ into the minipatch, because a tiny fraction of true signal features might still be outside of $\mathcal{A}$ due to noise and correlation among features. Intuitively, the ever-changing active feature set $\mathcal{A}$ and $\{\gamma^{(k)}\}$ work together to adaptively strike a balance between exploitation and exploration of the input feature space, thus further enhancing computational efficiency and feature selection accuracy (e.g. Figure \ref{fig:adaptive}).

\subsection{Theoretical Properties}

Suppose we observe training data $(\mathbf{y}, \mathbf{X})\in\mathbb{R}^N\times \mathbb{R}^{N\times M}$ that have $N$ observations and $M$ features. Let $S\subset \{1,\hdots,M\}$ denote the true feature support set, which is the set of indices corresponding to the underlying true signal features. And let $S^c=\{1,\hdots,M\}\setminus S$ be the complement set that contains indices corresponding to the noise features. Both $S$ and $S^c$ are typically unknown. One important goal of feature selection is to infer $S$ from the observed data.

\vspace{0.5em}
Following the notations in Algorithm \ref{alg:stamps}, let $\{(I_t, F_t)\}$ denote the collection of all possible uniformly random minipatches of a fixed size and let $(I, F) \in \{(I_t, F_t)\}$ represent an arbitrary minipatch among them. The estimated feature support on this minipatch by an arbitrary base selector is denoted as $\hat{\mathcal{S}}\subseteq F$. For a feature $j\in\{1,\hdots,M\}$, the probability of being in the estimated feature support given it is in the minipatch can be written as $\mathbb{P}[j\in\hat{\mathcal{S}} \:|\: j\in F]$. Lastly, $\pi_{\text{thr}}\in(0,1)$ is the cut-off defined in Algorithm \ref{alg:stamps}. Here, we provide some insights into the theoretical properties of the STAMPS method. The proof is found in the Appendix.

\vspace{1em}
\noindent \textbf{Assumption (A1)} Each feature $j\in \{1,\hdots,M\}$ is sampled at least once during the STAMPS procedure.

\vspace{1em}
\noindent \textbf{Assumption (A2)} For any noise feature $j\in S^c$, assume $\mathbb{P}[j\in\hat{\mathcal{S}} \:|\: j\in F]\leq \alpha \pi_{\text{thr}}/|S^c|$.

\vspace{1em}
\noindent \textbf{Theorem 1} \textit{Assume that Assumptions (\textbf{\emph{A1-2}}) hold. Then as $K\to\infty$, STAMPS controls the familywise error rate (FWER) at level $\alpha\in(0,1)$:}
\begin{align*}
    \mathbb{P}[|\hat{S}^{\text{stable}}\cap S^c|\geq 1] \leq \alpha
\end{align*}

A simple sampling strategy that satisfies Assumption (\textbf{A1}) is to sample non-overlapping blocks of features sequentially such that all features are sampled exactly once before proceeding with uniform random sampling. In Assumption (\textbf{A2}), the probability $\mathbb{P}$ is taken with respect to the randomness of minipatches. Intuitively, Assumption (\textbf{A2}) requires the base selector to have a sufficiently low probability of selecting noise features on random minipatches. Putting everything together, under Assumptions (\textbf{A1-2}), Theorem 1 provides error control showing that the probability of STAMPS mistakenly selecting at least one noise feature is at most $\alpha$. This result is noteworthy because it is a model-agnostic analysis of STAMPS in a general setting where we do not impose any restrictions on the distribution of the underlying data, or the specific type of base feature selector, or the type of machine learning task.

\vspace{0.5em}
While one could argue that Assumption (\textbf{A2}) might be a bit stringent in practice, it is actually slightly looser than the exchangeability assumptions of stability selection \citep{meinshausen2010stability} which establishes false discovery rate control in the linear regression setting or the irrepresentable condition \citep{zhao2006consistency} required for the Lasso procedure to be selection consistent. Given the strong empirical performance of our methods, we believe that our assumptions could be relaxed, perhaps markedly so by studying specific base feature selectors. Additionally, the strong empirical performance of AdaSTAMPS suggests error rate control, but establishing this is beyond the scope of this paper and left for future work.

\subsection{Practical Considerations}

\begin{figure}[!h]
    \centering
    \includegraphics[width=1\linewidth]{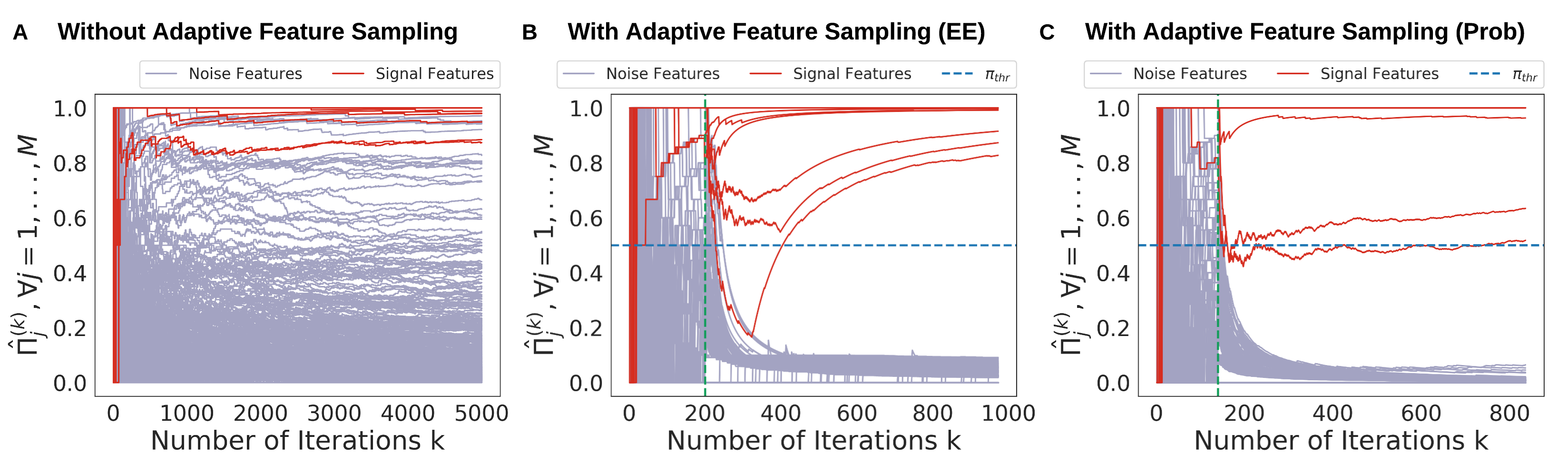}
    \caption{{\small \textit{Performance of our approaches on a challenging feature selection task in the regression setting with highly correlated features (as described in the Appendix). (A) Feature selection frequencies by STAMPS versus number of iterations; (B) Feature selection frequencies by AdaSTAMPS (EE) versus number of iterations with the green dotted line denoting the end of the burn-in stage; (C) Feature selection frequencies by AdaSTAMPS (Prob) versus number of iterations. This reveals that well-designed adaptive feature sampling schemes help AdaSTAMPS correctly select all true signal features (red) while discarding all noise features (gray) within much fewer iterations, thus further boosting both feature selection accuracy and computational efficiency in practice.}}}
    \label{fig:adaptive}
    \end{figure}

Our STAMPS and AdaSTAMPS are general meta-algorithms and they have three tuning hyperparameters: minipatch size ($n$ and $m$) and threshold $\pi_{\text{thr}}$. In general, our methods are quite robust to the choice of these hyperparameters. In our empirical studies in Sec. 3, we use a default threshold $\pi_{\text{thr}}=0.5$, which generally works well in practice. We also propose a data-driven way to choose $\pi_{\text{thr}}$ as detailed in the Appendix. For choosing the minipatch size, we have empirical studies in the Appendix investigating how feature selection accuracy and runtime vary for various $n$ and $m$ values. In general, we found taking $m$ to well exceed the number of true features (e.g. 3-10 times the number of true features) and then picking $n$ relative to $m$ so that it well exceeds the sample complexity of the base feature selector used in step 2 of Algorithm \ref{alg:stamps} works well in practice. Our empirical results also reveal that our meta-algorithms have stable performance for any sensible range of $n$ and $m$.

\vspace{0.5em}
We also empirically illustrate that using adaptive feature sampling not only increases computational efficiency but also seems to further improve feature selection accuracy in practice. In Figure \ref{fig:adaptive}, we display selection frequencies of all features versus the number of iterations for STAMPS, AdaSTAMPS (EE), and AdaSTAMPS (Prob) on a challenging feature selection task. We see that both AdaSTAMPS (EE) and AdaSTAMPS (Prob) correctly identify all true signal features while discarding all noise features within $1000$ iterations. On the other hand, STAMPS without adaptive feature sampling converges slower and is unable to clearly separate a handful of true signal features from the noise features. Details of this empirical study are given in the Appendix.

\section{Empirical Studies}

In this section, we evaluate our proposed STAMPS and AdaSTAMPS approaches on both synthetic and real-world data sets. Because many people study the problem of feature selection in the linear regression setting and it is also easier to simulate synthetic data from linear models for comparisons with existing methods, we first focus on feature selection in the high-dimensional linear regression setting in Sec. 3.1. However, our proposed feature selection frameworks are not limited to either linear or regression settings. We showcase the superior performance of our AdaSTAMPS method on real-world classification problems in Sec. 3.2.

\subsection{Feature Selection in High-Dimensional Linear Regression}

We compare the feature selection accuracy as well as computational time of our proposed methods with several widely-used competitors including the Lasso \citep{tibshirani1996lasso}, Complementary Pairs Stability Selection (CPSS) \citep{shah2013cpss}, Randomized Lasso \citep{meinshausen2010stability}, univariate linear regression filter with False Discovery Rate (FDR) control via the Benjamini-Hochberg procedure, recursive feature elimination (RFE) \citep{guyon2002rfe}, and thresholded Ordinary Least Squares (OLS) \citep{giurcanu2016}. Our focus is on high-dimensional cases with $N \ll M$ because these settings are arguably the most difficult for feature selection. Additional results about feature interactions can be found in the Appendix. In this section, we consider three high-dimensional scenarios with the following data matrices $\mathbf{X}\in\mathbb{R}^{N\times M}$:

\begin{itemize}
    \item {\bf Scenario 1 (S1)} Autoregressive Toeplitz design: the $M=10000$-dimensional feature vector follows a $\mathcal{N}(\mathbf{0}, \vect{\Sigma})$ distribution, where $\Sigma_{ij}=\rho^{|i-j|}$ with $\rho=0.95$; sample size $N=5000$.
    \item {\bf Scenario 2 (S2)} Subset of The Cancer Genome Atlas (TCGA) data: this data set ($N=2834$, $M=10000$) contains 450K methylation status of the top $10000$ most variable CpGs for $2834$ patients. This scenario is even more challenging than S1 due to strong and complex correlation structure among CpGs. This data set was obtained via {\tt TCGA2STAT} \citep{tcga2stat}.
    \item {\bf Scenario 3 (S3)} Fullsize TCGA data: this data set ($N=2834$, $M=335897$) contains 450K methylation status of $335897$ CpGs with complete records for $2834$ patients.
\end{itemize}

Following standard simulation procedures \citep{meinshausen2010stability}, we create the true feature support set $S$ by randomly sampling $S\subset\{1,\hdots,M\}$ with cardinality $|S|=20$. The $M$-dimensional sparse coefficient vector $\vect{\beta}$ is generated by setting $\beta_j=0$ for all $j\notin S$ and $\beta_j$ is chosen independently and uniformly in $[-3/b, -2/b]\cup [2/b, 3/b]$ for all $j\in S$. Finally, the response vector $\mathbf{y}\in\mathbb{R}^N$ can be generated by $\mathbf{y} = \mathbf{X}\vect{\beta} + \vect{\epsilon}$ where the noise vector $(\epsilon_1,\hdots,\epsilon_N)$ is IID $\mathcal{N}(0,1)$. For S1 and S2, we vary the positive real number $b$ to attain various signal-to-noise ratio (SNR) levels. For S3, SNR is fixed at $5$.

\vspace{0.5em}
For each method, commonly-used data-driven model selection procedures are applied to choose tuning hyperparameters and produce a data-driven estimate of the true feature support. In addition, oracle model selection that assumes knowledge of cardinality of the true support $|S|$ is employed with each method, resulting in an oracle estimate of the true feature support. Detailed descriptions of these data-driven and oracle model selection procedures are given in the Appendix. We evaluate the feature selection accuracy of various methods using the F1 Score, which takes values between $0$ and $1$ with $1$ indicating perfect match between the final estimated feature support and the true feature support $S$. For every method, we compute F1 Scores for both data-driven and oracle estimates.

\vspace{0.5em}
We use {\tt Scikit-learn} \citep{scikit-learn} for all Lasso-based methods, univariate filter, and RFE with Ridge estimator. Due to lack of existing software, we faithfully implemented the thresholded OLS as in \citet{giurcanu2016}. For simplicity, our methods (i.e. STAMPS, AdaSTAMPS (EE), and AdaSTAMPS (Prob)) are employed with the same thresholded OLS as base selector in step 2 of Algorithm \ref{alg:stamps} using default settings as detailed in the Appendix. All comparisons were run on a VM with $10$ Intel (Cascade Lake) vCPUs and $220$ GB of memory.

\begin{figure}[h]
    \centering
    \includegraphics[width=0.85\linewidth]{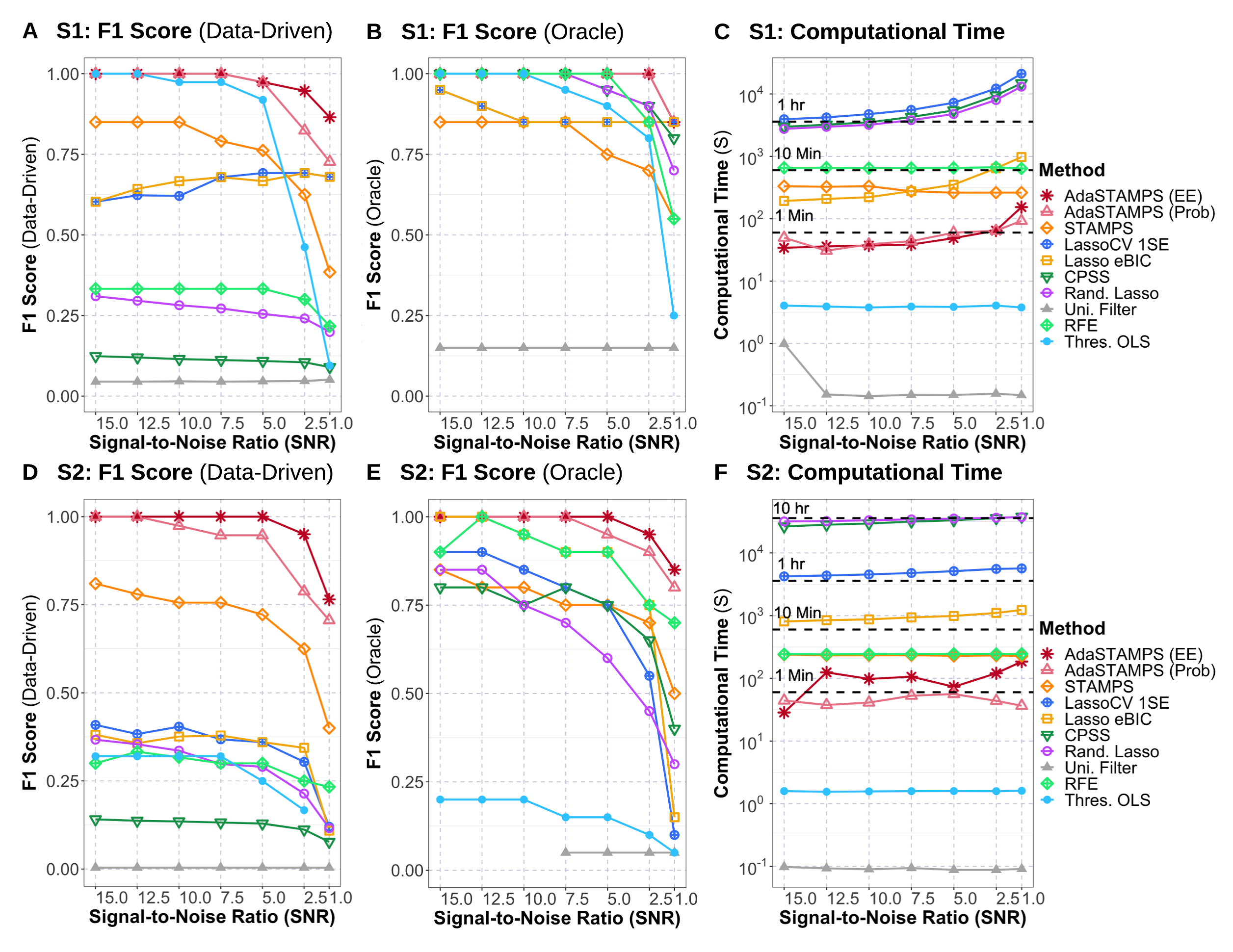}
    \caption{{\small \textit{Feature selection accuracy (F1 Score) and runtime from simulation Scenario 1 (S1) and Scenario 2 (S2). (A, D) F1 Score for data-driven feature support estimates; (B, E) F1 Score for oracle feature support estimates; (C, F) Computational time in seconds. The fastest runtime is reported for every method with parallelism enabled for methods (LassoCV 1SE, Lasso eBIC, CPSS, Rand. Lasso, RFE) wherever possible. Our AdaSTAMPS dominates all competitors in terms of feature selection accuracy in both scenarios. Moreover, the runtime of our AdaSTAMPS is the best of any method with comparable feature selection accuracy.}}}
    \label{fig:smallcomp}
    \end{figure}

\vspace{0.5em}
The feature selection accuracy and computational time of various methods from scenarios S1 and S2 are shown in Figure \ref{fig:smallcomp}. We see that our AdaSTAMPS consistently dominates all competing methods in terms of feature selection accuracy in both scenarios. While some competitors such as RFE and the Lasso achieve comparable accuracy to AdaSTAMPS with oracle model selection at some SNR levels (Figure \ref{fig:smallcomp}B, \ref{fig:smallcomp}E), their performance degrades drastically when their tuning hyperparameters are chosen via widely-used data-driven ways (Figure \ref{fig:smallcomp}A, \ref{fig:smallcomp}D). On the other hand, performance of AdaSTAMPS doesn't deteriorate as much across data-driven and oracle model selection. We also point out that our AdaSTAMPS can boost the accuracy of any existing feature selection strategy - AdaSTAMPS with thresholded OLS as base selector outperforms thresholded OLS itself by a large margin, especially in low SNR settings and in the more challenging scenario S2. Furthermore, AdaSTAMPS is computationally the fastest of any method with comparable accuracy, with runtime on the order of seconds to minutes as opposed to hours for many competing methods.

\vspace{0.5em}
The results for the most challenging scenario S3 are summarized in Table \ref{tcga-table}. High-dimensional data of such scale is regularly observed in areas such as genetics and neuroimaging. We see that our AdaSTAMPS (EE) perfectly recovers the true feature support across both data-driven and oracle model selection. In addition, STAMPS and AdaSTAMPS (Prob) also perform favorably. On the other hand, all competing methods attain rather low feature selection accuracy even with oracle selection. Clearly, AdaSTAMPS (EE) is more capable of accurately identifying the true signal features even in high-correlation, high-dimensional settings, thus leading to more reliable model interpretation. In the meantime, AdaSTAMPS (EE) takes less than $20$ minutes to run, rather than the $14$ hours required to solve Randomized Lasso that achieves the next best data-driven F1 Score among competing methods.

\columnalignment{tag1}{r}
\begin{table*}[h]
  \caption{Results of Various Methods for Simulation Scenario 3 (S3). For each method, the number of selected features determined by data-driven procedures, the data-driven F1 Score, the oracle F1 Score, and computational time in minutes are reported. The method with the best accuracy is bold-faced. Note that the fastest runtime is reported for every method with parallelism enabled for methods ($*$) wherever possible. Our AdaSTAMPS dominates all competing methods in terms of feature selection accuracy. }
  \label{tcga-table}
  \scalebox{0.9}{\parbox{1\linewidth}{%
  \centering
  \begin{tabular}{l>{\collectcell\columnalign}c<{\endcollectcell}>{\collectcell\columnalign}c<{\endcollectcell}>{\collectcell\columnalign}c<{\endcollectcell}>{\collectcell\columnalign}c<{\endcollectcell}}
    \toprule
    \multicolumn{1}{c}{Method} &\multicolumn{1}{c}{ \# Selected Features} & \multicolumn{1}{c}{F1 (Data-Driven)} & \multicolumn{1}{c}{F1 (Oracle)} & \multicolumn{1}{c}{Time (Minute)} \\
    \midrule
    AdaSTAMPS (EE) & \textbf{20} & \textbf{1.000} & \textbf{1.000} & $19.87$ \\
    AdaSTAMPS (Prob) & 47 & 0.597 & 0.750 & $52.10$ \\
    STAMPS & 237 & 0.117 & 0.750 & $74.70$ \\
    LassoCV 1SE & 194 & 0.112 & 0.600 & $176.60^{*}$ \\
    Lasso eBIC & 46 & 0.091 & 0.100 & $61.59^{*}$ \\
    CPSS & 81 & 0.238 & 0.500 & $716.94^{*}$ \\
    Rand. Lasso & 9 & 0.276 & 0.450 & $838.53^{*}$ \\
    Uni. Filter & 277261 & 0.000 & 0.050 & $0.05$ \\
    RFE & 37035 & 0.001 & 0.350 & $97.82^{*}$ \\
    Thres. OLS & 4 & 0.083 & 0.100 & $1.01$ \\
    \bottomrule
  \end{tabular}
  }}
\par
\end{table*}

\vspace{0.5em}
Overall, we empirically demonstrate the effectiveness of our proposed approaches in selecting statistically accurate features with much improved robustness and computational efficiency, even in challenging high-noise, high-correlation scenarios. Even though STAMPS does not perform as well as AdaSTAMPS, STAMPS still performs better than a lot of the competing methods, especially in terms of data-driven F1 Score (e.g. Figure \ref{fig:smallcomp}A, \ref{fig:smallcomp}D). In the meantime, our AdaSTAMPS (both (EE) and (Prob)) consistently outperform competing methods in terms of feature selection accuracy while being computationally the fastest of any method with comparable accuracy. This suggests that our flexible AdaSTAMPS approach could be a double-win both in terms of statistical accuracy and computation in practice. Intuitively speaking, the strong empirical performance of our proposed methods can be attributed to the following reasons: i) the use of tiny minpatches yields major computational advantages and our adaptive sampling schemes further boost efficiency; ii) the repeated subsampling of features breaks up strong correlations among features that could otherwise derail many existing feature selection strategies; and iii) the final aggregation of selection events over all minipatches reduces variance in a similar manner to that of random forests and hence ensures strong performance even in high-noise-low-signal settings.

\subsection{Real-World Classification Data}

We now study the performance of our method when employed as a dimensionality-reduction preprocessing step for downstream classification tasks on several real-world data sets - CIFAR10 \citep{Krizhevsky09learningmultiple}, Gisette \citep{gisette2004}, and Fashion MNIST \citep{Xiao2017FashionMNISTAN}. For CIFAR10, we consider two classes (\textit{truck}, \textit{automobile}) with $N=12000$ images each having $M=3072$ features. Gisette is a binary classification data set ($N=7000$, $M=5000$) . For Fashion MNIST, we consider three classes (\textit{T-shirt}, \textit{coat}, \textit{shirt}) with $N=18000$ and $M=784$.  We split each data set into $70\%$ training and $30\%$ test data via stratified sampling.

\begin{figure*}[!ht]
    \centering
    \includegraphics[width=0.85\linewidth]{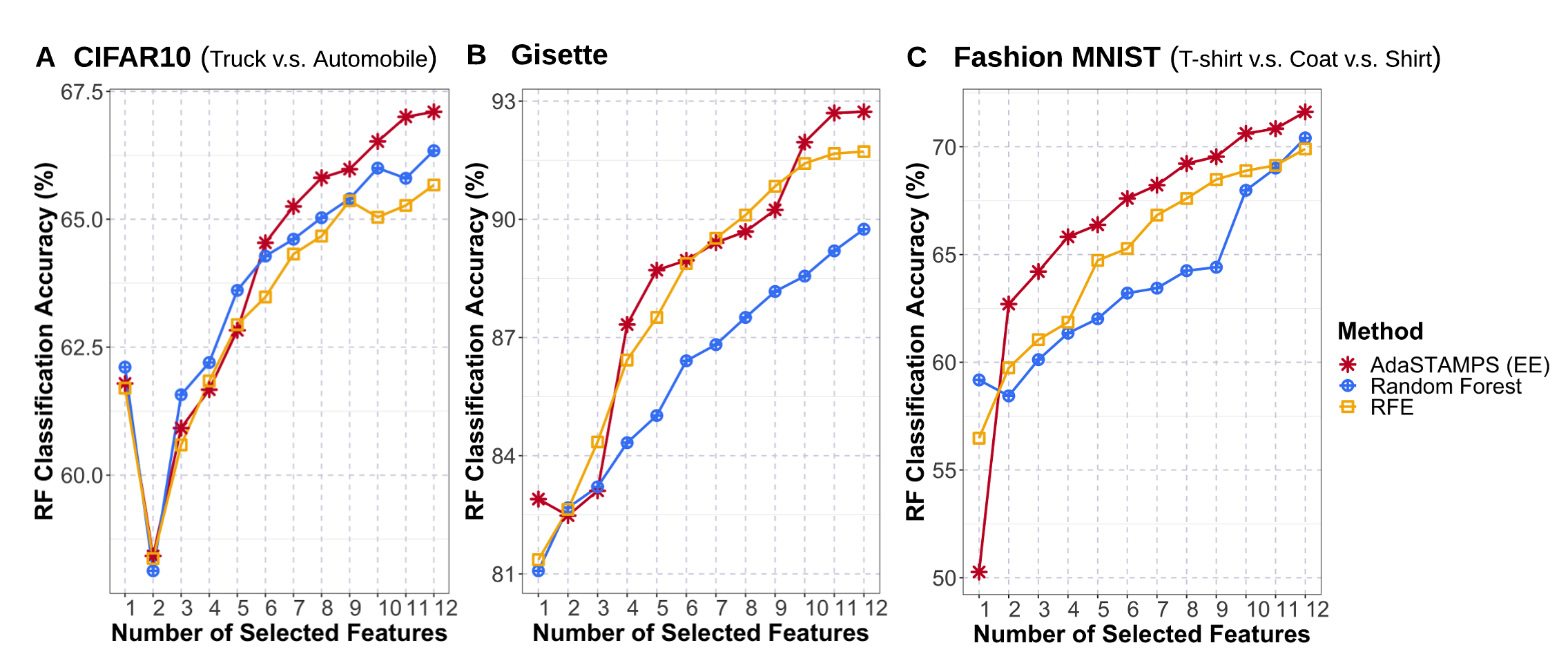}
    \caption{{\small \textit{Results for the real-world classification data. The top features selected by each feature selection method are used to train new RF classifiers separately for downstream classification tasks. (A, B, C) Test accuracy of these downstream RF classifiers (averaged over $5$ random training/test splits) versus number of selected top features that are used in training the downstream classifiers. Overall, the features selected by our AdaSTAMPS (EE) lead to better classification accuracy on test sets, especially as the number of selected features increases.}}}
    \label{fig:classification}
    \end{figure*}

\vspace{0.5em}
For every training data set, our AdaSTAMPS (EE) fits Random Forest (RF) classifiers as base selectors to many random minipatches and keeps $10$ features with the highest Gini importance from every minipatch. Features are then ranked based on their selection frequencies from AdaSTAMPS (EE), which are computed by taking an ensemble of selection results over these minipatches. For comparisons, we also fit a RF classifier to the full training data and obtain ranks for features based on the RF Gini importance. Moreover, we obtain another set of ranks for features using a RFE with RF as base estimator on the full training data. As the true feature support is unknown in these real-world data, performance of each feature selection method is measured by training a new downstream RF classifier using only its top ranked $J$ features and then computing classification accuracy on the test data. In all cases, we use the default RF classifier from {\tt Scikit-learn}. Figure \ref{fig:classification} shows the average test accuracy of the downstream RF classifiers over $5$ random training/test splits versus number of selected features that are used in training these downstream classifiers.

\vspace{0.5em}
From Figure \ref{fig:classification}, we see that AdaSTAMPS (EE) mostly outperforms the competitors because the features selected by AdaSTAMPS (EE) are generally more informative and eventually lead to better test accuracy in all three data sets. Similar to observations from Sec. 3.1, our AdaSTAMPS framework can boost the feature selection performance of any existing selection strategy: AdaSTAMPS (EE) with RF as inner base feature selector outperforms RF itself, sometimes by a large margin as in Figure \ref{fig:classification}B.

\section{Conclusion}

We have developed general meta-algorithms for feature selection named STAMPS and AdaSTAMPS that leverage random minipatches as well as adaptive feature sampling schemes. We empirically demonstrate that our approaches, especially AdaSTAMPS, have superior feature selection performance and computational efficiency for a variety of synthetic and real-world scenarios. The main advantage of our methods is that these are practical solutions that we can employ with a wide variety of existing learning algorithms and feature selection strategies to identify statistically accurate features in a computationally efficient manner, thus enhancing interpretability of machine learning models.

\vspace{0.5em}
In addition to demonstrating the practical effectiveness of our proposed methods, we have provided theoretical insights on STAMPS in terms of model-agnostic familywise error rate control. In future work, we plan on working with specific base selector methods to establish stronger theoretical results and developing theoretical frameworks for analyzing the more sophisticated AdaSTAMPS method. We also would like to investigate distributed, memory-efficient implementation of our methods.

\section*{Acknowledgements}

The authors acknowledge support from NSF DMS-1554821, NSF NeuroNex-1707400, and NIH 1R01GM140468.

\newpage


\begin{center}
{\bf \LARGE Feature Selection for Huge Data via Minipatch Learning:\\
Supplementary Materials}
\bigskip

{\large Tianyi Yao and Genevera I. Allen}
\end{center}

\renewcommand\thefigure{\thesection{}}
\renewcommand\thetable{\thesection.\arabic{table}}
\renewcommand\thesection{\Alph{section}}
\renewcommand\thesubsection{\thesection.\arabic{subsection}}

\setcounter{section}{0}
\section{Probabilistic Adaptive Feature Sampling Scheme}

In Algorithm 3, we give our probabilistic adaptive feature sampling scheme. Here, $k$ is the iteration counter, $E$ denotes the total number of burn-in epochs. We use the shorthand AdaSTAMPS (Prob) to denote the specific instance of AdaSTAMPS that employs Algorithm 3 as its adaptive feature sampling scheme.

Similar to our exploitation and exploration scheme (Algorithm 2 in the main paper), AdaSTAMPS (Prob) first explores the entire input feature space during the burn-in stage. After $E$ burn-in epochs, Algorithm 3 transitions to the adaptive stage during which it adaptively updates the sampling probabilities for all features. Specifically, features with higher selection frequencies, which are likely to be the true signal features, have relatively higher probabilities of being sampled into subsequent minipatches. However, since this is a probabilistic sampling scheme, Algorithm 3 still adaptively explores the remaining feature space because many features with relatively lower selection frequencies still have non-zero probabilities of being sampled into minipatches.

{\centering
\setcounter{algocf}{2}
\begin{algorithm}[H]
\label{alg:adastamps-probabilistic}
\caption{Adaptive Feature Sampling Scheme: Probabilistic (Prob)}
\SetAlgoLined
\KwIn{$k$, $M$, $m$, $E$, $\{\hat{\Pi}_j^{(k-1)}\}_{j=1}^{M}$.}
\vspace{5pt}
\textbf{Initialize:} $G=\ceil{\frac{M}{m}}$, $\mathcal{J}=\{1,\hdots,M\}$\;
\vspace{5pt}
    \uIf(\tcp*[h]{Burn-in stage}){$k \leq E\cdot G$}{
        \vspace{5pt}
        \uIf(\tcp*[h]{Start of a burn-in epoch}){$\text{mod}_{G}(k)=1$}{
            \vspace{3pt}
            1) Randomly reshuffle feature index set $\mathcal{J}$ and partition into disjoint sets $\{\mathcal{J}_g\}_{g=0}^{G-1}$\;
        }
        \vspace{3pt}
        1) Set $F_k=\mathcal{J}_{\text{mod}_{G}(k)}$\;
    }
    \Else(\tcp*[h]{Adaptive stage}){
        \vspace{3pt}
        1) Update feature sampling probabilities using information from previous iterations:

        \begin{align*}
            \hat{p}^{(k)}_j = \frac{\hat{\Pi}^{(k-1)}_j}{\sum_{j=1}^M\hat{\Pi}^{(k-1)}_j} \:\:\:\:\text{for}\:\:\:\:j\in \{1,\hdots,M\}
        \end{align*}

        2) Sample $m$ features $F_k \subset \{1,\hdots,M\}$ without replacement according to the updated feature sampling probabilities $\{\hat{p}^{(k)}_j\}_{j=1}^M$\;

    }
{\bf Output:} $F_k$.
\end{algorithm}
}

\section{Proof of Theorem 1}
In this section, we present the detailed proof of Theorem 1. We restate our theoretical statements from the main paper here for convenience.

\vspace{1em}
\noindent Following the notations in Algorithm 1, let $\{(I_t, F_t)\}$ denote the collection of all possible uniformly random minipatches of a fixed size and let $(I, F) \in \{(I_t, F_t)\}$ represent an arbitrary minipatch among them. The estimated feature support on this minipatch by an arbitrary base selector is denoted as $\hat{\mathcal{S}}\subseteq F$. For a feature $j\in\{1,\hdots,M\}$, the probability of being in the estimated feature support given it is in the minipatch can be written as $\mathbb{P}[j\in\hat{\mathcal{S}} \:|\: j\in F]$. Lastly, $\pi_{\text{thr}}\in(0,1)$ is the cut-off defined in Algorithm 1.


\vspace{1em}
\noindent \textbf{Assumption (A1)} Each feature $j\in \{1,\hdots,M\}$ is sampled at least once during the STAMPS procedure.



\vspace{1em}
\noindent \textbf{Assumption (A2)} For any noise feature $j\in S^c$, assume $\mathbb{P}[j\in\hat{\mathcal{S}} \:|\: j\in F]\leq \alpha \pi_{\text{thr}}/|S^c|$.

\vspace{1em}
\noindent \textbf{Theorem 1} \textit{Assume that Assumptions (\textbf{\emph{A1-2}}) hold. Then as $K\to\infty$, STAMPS controls the familywise error rate (FWER) at level $\alpha\in(0,1)$:}
\begin{align*}
    \mathbb{P}[|\hat{S}^{\text{stable}}\cap S^c|\geq 1] \leq \alpha
\end{align*}

\noindent \textit{Proof:} By Markov inequality, we have
\begin{align*}
    \mathbb{P}[|\hat{S}^{\text{stable}}\cap S^c|\geq 1] &\leq \frac{\mathbb{E}\big[|\hat{S}^{\text{stable}}\cap S^c|\big]}{1} \\
    &= \mathbb{E}\big[\sum_{j\in S^c}\mathds{1}(j \in \hat{S}^{\text{stable}})\big] \\
    &= \mathbb{E}\big[\sum_{j\in S^c}\mathds{1}(\hat{\Pi}_j^{(K)} \geq \pi_{\text{thr}})\big] \\
    &= \sum_{j\in S^c} \mathbb{P}[\hat{\Pi}_j^{(K)} \geq \pi_{\text{thr}}] \numberthis \label{proof:1}
\end{align*}
By definition, $\hat{\Pi}_j^{(K)}\geq 0$ for all $j=1,2,\hdots,M$ and $\pi_{\text{thr}} \in (0,1)$. Using Markov inequality, for any noise feature $j\in S^c$, we have
\begin{align*}
        \mathbb{P}[\hat{\Pi}_j^{(K)} \geq \pi_{\text{thr}}] &\leq \frac{\mathbb{E}[\hat{\Pi}^{(K)}_j]}{\pi_{\text{thr}}} \numberthis \label{proof:2}
\end{align*}
Now we need to evaluate $\mathbb{E}[\hat{\Pi}^{(K)}_j]$. As defined in Algorithm 1, the selection frequency for any noise feature $j\in S^c$ is
\begin{align*}
    \hat{\Pi}^{(K)}_j &= \frac{\sum_{t=1}^K\mathds{1}(j\in F_t, j\in \hat{\mathcal{S}}_t)}{\text{max}(1,\sum_{t=1}^K \mathds{1}(j\in F_t))} \\
    &= \frac{\sum_{t=1}^K\mathds{1}(j\in F_t, j\in \hat{\mathcal{S}}_t)}{\sum_{t=1}^K \mathds{1}(j\in F_t)} \:\:\:\:\:\: (\text{by Assumption (\textbf{A1})}) \\
    &= \frac{\frac{1}{K}\sum_{t=1}^K\mathds{1}(j\in F_t, j\in \hat{\mathcal{S}}_t)}{\frac{1}{K}\sum_{t=1}^K \mathds{1}(j\in F_t)} \\
    &\convp \frac{\mathbb{E}[\mathds{1}(j\in F, j\in \hat{\mathcal{S}})]}{\mathbb{E}[\mathds{1}(j\in F)]} \:\:\:\:\:\: (\text{by the Law of Large Numbers as}\:\:K\to\infty) \\
    &= \frac{\mathbb{P}[j\in F, j\in \hat{\mathcal{S}}]}{\mathbb{P}[j\in F]} \\
    &= \mathbb{P}[j\in \hat{\mathcal{S}} \:|\: j\in F] \\
\end{align*}
Then for any given noise feature $j\in S^c$, we have
\begin{align*}
    \mathbb{E}[\hat{\Pi}^{(K)}_j] &= \mathbb{E}\big[\mathbb{P}[j\in \hat{\mathcal{S}} \:|\: j\in F]\big] \\
    &\leq \mathbb{E}\Big[\frac{\alpha \pi_{\text{thr}}}{|S^c|}\Big] \:\:\:\:\:\: (\text{by Assumption (\textbf{A2})}) \\
    &= \frac{\alpha \pi_{\text{thr}}}{|S^c|} \numberthis \label{proof:3}
\end{align*}
Then the upper bound in \eqref{proof:2} becomes
\begin{align*}
        \mathbb{P}[\hat{\Pi}_j^{(K)} \geq \pi_{\text{thr}}] &\leq \frac{\mathbb{E}[\hat{\Pi}^{(K)}_j]}{\pi_{\text{thr}}} \\
        &\leq \frac{1}{\pi_{\text{thr}}}\frac{\alpha \pi_{\text{thr}}}{|S^c|} \\
        &= \frac{\alpha}{|S^c|} \numberthis \label{proof:4}
\end{align*}
and combining with \eqref{proof:1}, we have
\begin{align*}
    \mathbb{P}[|\hat{S}^{\text{stable}}\cap S^c|\geq 1] &\leq \sum_{j\in S^c} \mathbb{P}[\hat{\Pi}_j^{(K)} \geq \pi_{\text{thr}}] \\
    &\leq \sum_{j\in S^c} \frac{\alpha}{|S^c|} \\
    &= \alpha \numberthis \label{proof:5}
\end{align*}

\newpage
\section{More on Practical Considerations}

\subsection{Data-Driven Rule to Choose $\pi_{\text{thr}}$}

While there are many possible data-driven approaches to choose the user-specific threshold $\pi_{\text{thr}}\in(0,1)$, we propose a very simple one in Algorithm 4 that we found to be effective in practice. Here, $\{\hat{\Pi}_j^{(K)}\}_{j=1}^M$ is the set of feature selection frequencies from the last iteration of STAMPS or AdaSTAMPS, and $\hat{\text{SD}}$ denotes the sample standard deviation. Intuitively, Algorithm 4 tries to find all gaps (local minima of density) among $\{\hat{\Pi}_j^{(K)}\}_{j=1}^M$ in order to choose a threshold that separates features with relatively high selection frequencies from those with low selection frequencies. Algorithm 4 does so by first fitting a Gaussian kernel density estimator (KDE) to the selection frequencies and then setting $\pi_{\text{thr}}$ to the smallest local minima, if any.

{\centering
\scalebox{1}{\parbox{1\linewidth}{%
\setcounter{algocf}{3}
\begin{algorithm}[H]
\label{kde-rule}
\caption{KDE-Based Data-Driven Rule to Choose $\pi_{\text{thr}}$}
\SetAlgoLined
\KwIn{$\{\hat{\Pi}_j^{(K)}\}_{j=1}^M$.}
\vspace{5pt}

    1) Fit a Gaussian KDE to $\{\hat{\Pi}_j^{(K)}\}_{j=1}^M$:
    \begin{align*}
        \hat{f}_h(x) = \frac{1}{M}\sum_{j=1}^M \exp{\Big(\frac{(x-\hat{\Pi}_j^{(K)})^2}{2h^2}\Big)}, \forall x \in [0,1]
    \end{align*}
    where the bandwidth of the Gaussian kernel $h=\hat{\text{SD}}(\{\hat{\Pi}_1^{(K)},\hat{\Pi}_2^{(K)},\hdots, \hat{\Pi}_M^{(K)}\})$\;
    \vspace{5pt}
    2) Use method of inflection to find all local minima:
    \begin{align*}
        \mathcal{D}=\{x\in[0,1]\:\:\text{s.t.}\:\:\text{sign}(\hat{f}_h(x)-\hat{f}_h(x-\epsilon))=-1, \text{sign}(\hat{f}_h(x+\epsilon)-\hat{f}_h(x))=1\}
    \end{align*}
    for any $\epsilon>0$\;
    \vspace{5pt}
    \uIf(\tcp*[h]{Find at least one local minima}){$|\mathcal{D}|\geq 1$}{
    \vspace{5pt}
    Set $\pi_{\text{thr}}=\text{min}(\mathcal{D})$\;
    }
    \Else(\tcp*[h]{No local minima is found}){
    \vspace{5pt}
    Set $\pi_{\text{thr}}=0.5$\;
    }
{\bf Output:} $\pi_{\text{thr}}$.
\end{algorithm}
}}
\par
}

\subsection{Additional Empirical Studies to Investigate Effects of Minipatch Size}
Here, we empirically investigate how feature selection accuracy and runtime of our meta-algorithm vary for various minipatch sizes (i.e. $n$ and $m$ values). We take a subset of the TCGA data that contains 450K methylation status of the top $M=2000$ most variable CpGs for $N=2000$ patients as the data matrix $\mathbf{X}\in\mathbb{R}^{N\times M}$. Following similar procedures to Sec. 3.1 of the main paper, we create the true feature support $S$ by randomly sampling $S\subset\{1,\hdots,M\}$ with $|S|=10$. The $M$-dimensional sparse coefficient vector $\vect{\beta}$ is generated by setting $\beta_j=0$ for all $j\notin S$ and $\beta_j$ is chosen independently and uniformly in $[-3/b, -2/b]\cup [2/b, 3/b]$ for all $j\in S$. Finally, the response vector $\mathbf{y}\in\mathbb{R}^N$ can be generated by $\mathbf{y} = \mathbf{X}\vect{\beta} + \vect{\epsilon}$ where the noise vector $(\epsilon_1,\hdots,\epsilon_N)$ is IID $\mathcal{N}(0,1)$. The positive real number $b$ is chosen such that the SNR is fixed at $5$. This synthetic data set is also used for Figure 1 in the main paper.

\begin{figure}[H]
    \centering
    \includegraphics[width=1\linewidth]{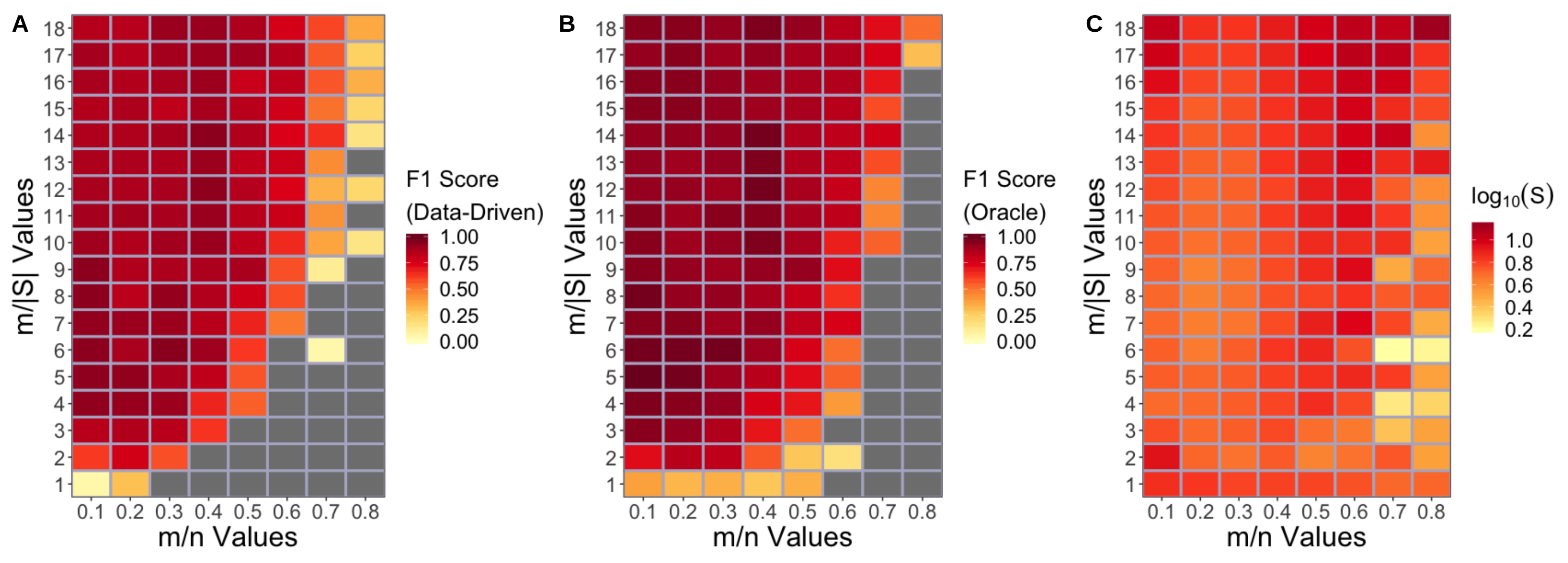}
    \caption{{\small \textit{We show how feature selection accuracy and computational time of AdaSTAMPS (EE) vary with minipatch sizes in terms of $m/|S|$ and $m/n$, where $|S|$ is the cardinality of the true feature support. F1 Scores and computational time are averaged over $5$ replicates. (A) Feature selection accuracy in terms of F1 Score with data-driven $\pi_{\text{thr}}$; (B) Feature selection accuracy in terms of F1 Score with oracle model selection; and (C) Computational time on $\log_{10}(\text{second})$ scale. Note that gray cells indicate NA's that result from poor selection performance. We see that our method has stable performance for a sensible range of $n$ and $m$.}}}
    \label{heatmap}
    \end{figure}

\vspace{0.5em}
To study how performance varies with minipatch size ($n$ and $m$), we run AdaSTAMPS (EE) using thresholded OLS as the inner base selector for various $n$ and $m$ values. Specifically, we take a sequence of $m$ values that are integer multiples of $|S|$ and then pick $n$ relative to $m$ such that $m/n \in \{0.1, 0.2, \hdots, 0.8\}$. Feature selection accuracy in terms of F1 Scores and computational time are reported for various $n$ and $m$ values in Figure \ref{heatmap}. We see that our method is quite robust to the choice of minipatch size for it has stable feature selection performance for a sensible range of $n$ and $m$. In general, we found taking $m$ to well exceed the number of true features (e.g. 3-10 times the number of true features $|S|$) and then picking $n$ relative to $m$ so that it well exceeds the sample complexity of the base feature selector used in the meta-algorithm works well in practice.

\subsection{Additional Simulation Experiment with Feature Interactions}
We have conducted additional simulation experiments for noisy 3-way interaction XOR data. Specifically, we generated data matrix $\mathbf{X}$ with $1000$ observations each having $20$ features. To simulate 3-way XOR feature interaction, the class labels $\mathbf{y}$ are determined by the interactions between 3 active features. We evaluate feature selection accuracy for identifying the true active features in terms of false positive rate (FPR), true positive rate (TPR), and the F1 score.

\vspace{0.5em}
As shown in Table \ref{xor-table}, our AdaSTAMPS that employs random forest as base selector successfully identifies all true active features while having the lowest FPR. STAMPS also performs favorably. On the other hand, univariate filter, recursive feature elimination that uses random forest as base learner, and random forest itself all failed to identify a single active feature while incorrectly selecting many noise features. This suggests that our proposed methods can indeed tackle XOR-type feature interactions much better than univariate filtering. In addition, we note that our proposed meta-algorithms are very general and they inherit the feature selecting properties of whichever base machine learning method and feature selector we use. Hence, if we use a base machine learning method that can handle interactions, then our meta-algorithms will adeptly select the right interaction features.

\begin{table}[!hb]
  \caption{Feature selection accuracy for identifying the true active features in XOR problem.}
  \label{xor-table}
  \centering
  \begin{tabular}{cccc}
    \toprule
    Method & FPR & TPR & F1 \\
    \midrule
    AdaSTAMPS (EE) & \textbf{0.059} & \textbf{1.000} & \textbf{0.857} \\
    STAMPS & 0.353 & \textbf{1.000} & 0.500 \\
    Uni. Filter & 0.176 & 0.000 & nan  \\
    RFE & 0.235 & 0.000 & nan  \\
    RF & 0.294 & 0.000 & nan  \\
    \bottomrule
  \end{tabular}
\end{table}

\section{Data-Driven and Oracle Model Selection Procedures for Various Methods in Empirical Studies}

In this section, we discuss details of data-driven and oracle model selection procedures that we use for our empirical studies in Sec. 3.1 of the main paper.

\vspace{0.5em}
For the Lasso, data-driven model selection approaches such as the $10$-fold cross-validation with 1 standard error (SE) rule \citep{hastieelements} and the extended BIC (eBIC) rule \citep{chen2008ebic} are employed to determine the optimal regularization parameter, resulting in $\lambda_{\text{1SE}}$ and $\lambda_{\text{eBIC}}$, respectively. For univariate filter, the data-driven rule keeps all features that are significantly associated with the response with FDR controlled at the default level of $0.05$. Additionally, 10-fold cross-validation is carried out for RFE to choose the best tuning hyperparameters and the Bonferroni procedure \citep{giurcanu2016} is used with thresholded OLS to determine the selected features in a data-driven manner. The user-specific threshold $\pi_{\text{thr}}$ is set to $0.5$ for STAMPS, AdaSTAMPS (EE), AdaSTAMPS (Prob), CPSS, and Randomized Lasso. Following the default settings in {\tt Scikit-learn} \citep{scikit-learn}, we use a geometric sequence $\Lambda \in (0.001\lambda_{\text{max}}, \lambda_{\text{max}})$ as the set of candidate regularization parameters for all Lasso-based methods, where $\lambda_{\text{max}}$ is the minimum amount of regularization such that all estimated coefficients become zero. $|\Lambda|$ is set to the default $100$ for simulation scenarios S1-S2 and to $10$ for scenario S3. For AdaSTAMPS (EE), we also use default settings with $E=10$ burn-in epochs and $\pi_{\text{active}}=0.1$. For AdaSTAMPS (Prob), $E$ is also set to $10$. Lastly, for STAMPS, AdaSTAMPS (EE), AdaSTAMPS (Prob), the stopping criterion parameters $\tau_{\text{u}}$ and $\tau_{\text{l}}$ are set to $60$ and $30$, respectively.

\vspace{0.5em}
Assuming knowledge of the number of true signal features $|S|$, oracle model selection is also applied to every method. Specifically, oracle model selection is applied to STAMPS, AdaSTAMPS (EE), AdaSTAMPS (Prob), CPSS, and Randomized Lasso by keeping the top $|S|$ features with the highest selection frequencies. For the Lasso, oracle selection keeps the top $|S|$ features with the largest magnitude in terms of coefficient estimates at $\lambda_{\text{1SE}}$ and $\lambda_{\text{eBIC}}$, respectively. For univariate filter, the top $|S|$ features with the smallest p-values are kept and the top $|S|$ features with the largest magnitude in terms of OLS coefficient estimates are selected for thresholded OLS. Furthermore, RFE is carried out until exactly $|S|$ features are left for oracle model selection.


\end{document}